\documentclass[letterpaper, 10pt, conference]{ieeeconf}  

\usepackage{mathtools}
\usepackage[pdftex,final]{graphicx}
\usepackage{units}
\usepackage[english]{babel}
\usepackage[T1]{fontenc}
\usepackage[latin1]{inputenc}
\usepackage[usenames]{color}
\usepackage{comment}
\usepackage{balance}
\usepackage{url}
\usepackage{todonotes}
\usepackage{xcolor}
\usepackage{tikz}

\usepackage{textpos}
\usepackage{setspace}
\usepackage[bookmarks=true]{hyperref}
\usepackage[all]{hypcap}

\makeatletter
\let\NAT@parse\undefined
\makeatother
\usepackage[numbers]{natbib}
                                                                                     
\IEEEoverridecommandlockouts                         
\renewcommand{\baselinestretch}{1.0}
\hypersetup{
    colorlinks,
    linkcolor={red!50!black},
    citecolor={blue!50!black},
    urlcolor={blue!80!black}
}

\makeatletter
\def\input@path{{../figures/}}
\makeatother
\graphicspath{{../figures/}}

\title{\LARGE \bf Probabilistic Object Tracking using a Range Camera}

\author{Manuel W\"{u}thrich$^{1}$, Peter Pastor$^{2}$, Mrinal Kalakrishnan$^{2}$, Jeannette Bohg$^{1}$ and Stefan Schaal$^{1,2}$%
\thanks{$^{1}$ Autonomous Motion Department at the Max-Planck-Institute for Intelligent Systems, T\"ubingen, Germany}%
\thanks{$^{2}$ Computational Learning and Motor Control lab at the University of Southern California, Los Angeles, CA, USA}}

\begin{document}

\maketitle
\thispagestyle{empty}
\pagestyle{empty}

\begin{abstract}
We address the problem of tracking the 6-DoF pose of an
object while it is being manipulated by a human or a robot. We use a 
dynamic Bayesian network to perform inference and
compute a posterior distribution over the current object
pose. Depending on whether a robot or a human manipulates the object,
we employ a process model with or without knowledge of control
inputs. Observations are obtained from a range camera. As opposed to previous object 
tracking methods, we explicitly model self-occlusions and occlusions from the environment, e.g, the human
or robotic hand. This leads to a strongly non-linear observation model
and additional dependencies in the Bayesian network. We employ a Rao-Blackwellised 
particle filter to compute an estimate of the object
pose at every time step. In a set of experiments, we demonstrate the ability of
our method to accurately and robustly track the object pose in real-time while it is being
manipulated by a human or a robot. 
\end{abstract}

\begin{textblock*}{100mm}(.\textwidth,-10.3cm)
 \begin{spacing}{0.8}
 {\fontsize{8pt}{2pt}\selectfont \sffamily
\noindent 2013 IEEE/RSJ International Conference on\\
Intelligent Robots and Systems (IROS)\\
November 3-7, 2013. Tokyo, Japan}
\end{spacing}
\end{textblock*}%


\section{INTRODUCTION}
Manipulation of objects is one of the remaining key challenges of robotics. In recent years, tremendous progress has been made
in the area of data-driven grasp synthesis~\cite{bohg:2013}. Given an object, the
goal is to infer a suitable grasp that adheres certain properties,
e.g. stability or functionality. In many cases, this grasp
is then performed in an open-loop manner without taking any feedback
into account, e.g. in~\citep{Saxena:06,Bohg2010362,DangA12}. This
approach can lead to a very poor success rate especially
in the presence of noisy and incomplete sensor data, inaccurate
models, or in a dynamic environment.
We have recently shown that the robustness of grasp execution can be
greatly increased by continuously taking
tactile sensor feedback into account~\cite{Pastor_2011}. This enables
the robot to adapt to unforeseen situations.

In this paper, we use visual sensing to continuously track the 6-DoF
pose of an object during manipulation, which could enable the robot
to adapt its actions according to the perceived state of the
environment. 
In contrast to our previous work~\cite{Pastor_2011}, such adaptation would not rely on being in contact with the object. 
Visual tracking is also crucial for
precision manipulation tasks such as drilling a hole or inserting a
key into a lock~\cite{Righetti13}. These tasks require precise alignment of a tool in the robot hand with objects in the environment.

We present a real-time marker-less object tracking algorithm as a basis for
these kinds of systems. We consider the movement of an object as a
stochastic process and model it in a dynamic Bayesian
network. We perform inference in this network to
compute a posterior distribution over the current object
pose. We follow the general paradigm of a Bayes filter~\cite{thrun} in
which we (i) {\em predict\/} the current object pose given the previous pose and then
(ii) {\em update\/} the prediction given an observation. For the first
step, we use a process model that can either be dependent on control
inputs (in case the robot is moving the object) or be based on the simple
assumption that the object pose will not change significantly in a short time
(in case the object is not being held by the robot).
\begin{figure}[t]
 \vspace{0.2cm}
 \centering
 \includegraphics[width=\linewidth]{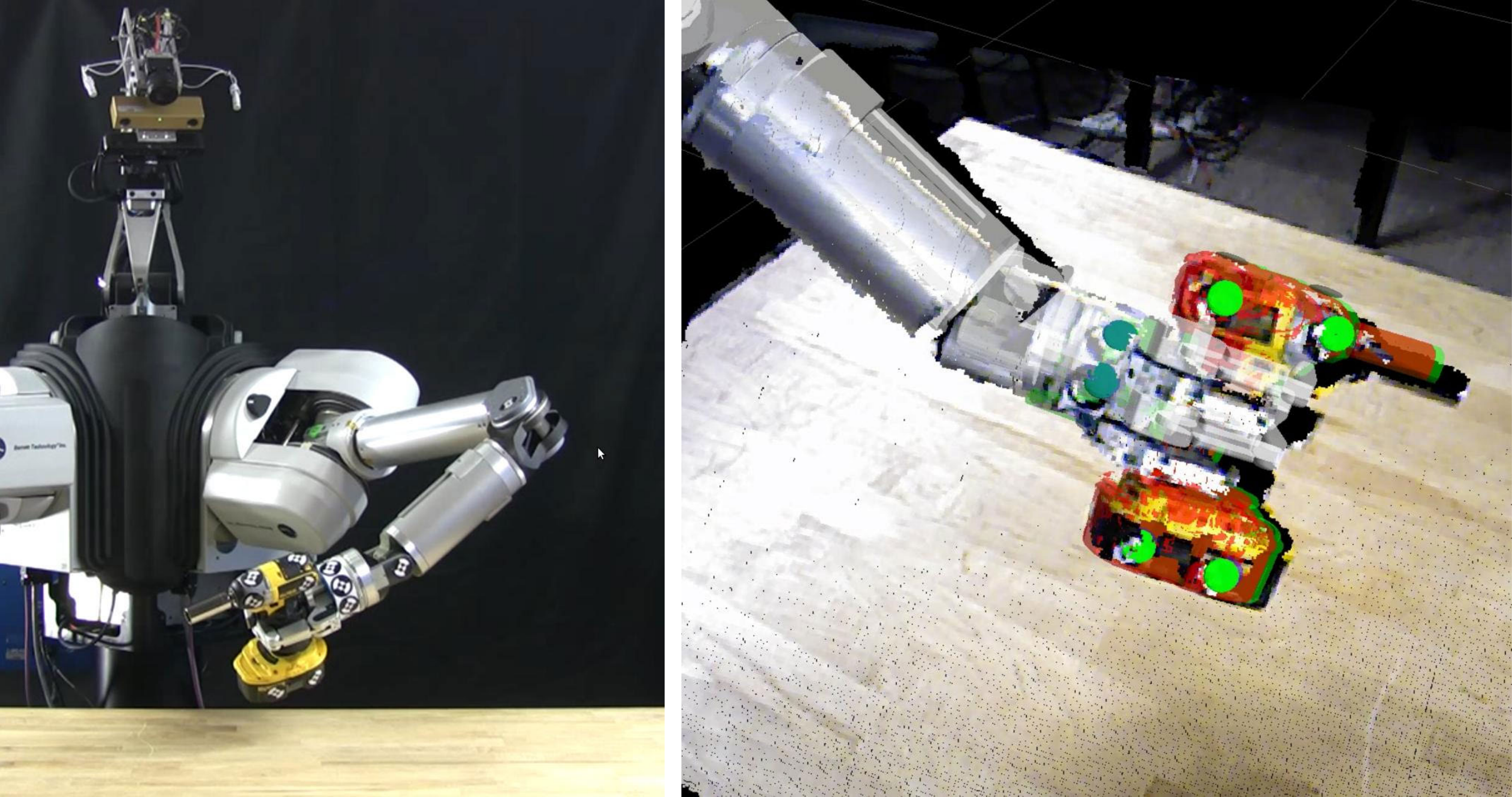}
 \vspace{-0.65cm}
 \caption{The robot is tracking an impact wrench (left image) in real-time using a depth sensor (red object model in the right image). Fiducial markers are used for baseline comparison only.}
 \label{fig:title}
 \vspace{-0.0cm}
\end{figure}
We use a range camera to obtain dense depth images of the scene.
For the update step, we develop
an observation model that, given the current estimate of the object
pose, determines the likelihood of the observed depth data. We explicitly model
self-occlusions and occlusions from the environment. Therefore, 
the algorithm becomes more robust to these effects. However, the observation
model is strongly non-linear and new dependencies are introduced in
the Bayes network. We therefore employ a Rao-Blackwellised particle filter \cite{doucet} to compute a posterior
distribution over the current object pose. 

The paper is structured as follows. In the next section, we review
existing approaches to object tracking, especially in the context
of robotic manipulation. We then briefly review the Bayes filter and  associated inference methods in
Sect.~\ref{sec:bayes_filter}. In Sect.~\ref{sec:obs} and~\ref{sec:process} the observation
and process models are derived. In Sect.~\ref{sec:filter} we formulate the proposed algorithm.
Experimental results are presented in Sect.~\ref{sec:exp}.
Finally, we conclude and present ideas for future work in Sect.~\ref{sec:conclusion}.

\section{RELATED WORK}
Existing approaches to real-time 6-DoF pose tracking can be divided into two groups according to the type of data used. 
The first group consists of algorithms which mainly rely on 2D images, e.g. \cite{harris90, kragic01, choi}. 
This kind of approaches may be sensitive to the amount of texture on the tracked object and in the scene as well as to lightning conditions. 
It has been shown that using depth data can add robustness to tasks such as SLAM~\cite{fusion}, object detection and 
recognition~\cite{lai_icra11b, hinterstoisser2012accv}.
Therefore it is worth exploring the second group of methods for 6 DoF pose tracking that rely on range data. In this
section, we review those which aim for real-time performance and use dense depth data as sensory input. 

\citet{ren} represent the object surface as the zero level in a 
level-set embedding function. Specifically, it is based on a 
truncated 3D distance map. 
The pose of the object is found by minimizing over the sum of all points
in the 3D point cloud back-projected into the object frame and
evaluated in the embedding function. Optimization is performed through the
Levenberg-Marquardt algorithm.  The GPU implementation of this
approach performs in real-time. Through the choice of a robust
estimator and a robust variant of the 3D distance map it seems
to work well even in the presence of heavy occlusions. 
There are however no experiments provided where an occluding object is in contact with the tracked object. 
It is not clear how the algorithm would perform in that situation since points from the occluding object 
might be mistaken for the tracked object.
Our method, in contrast, is formulated in a Bayesian framework that
allows us to fuse different sources of information in a probabilistic
manner, we fuse for example the knowledge of the control inputs with the depth measurements in order to estimate the
pose of the object. Furthermore, we explicitly model
occlusions in this framework and finally, 
we obtain a posterior distribution over the pose instead of just a point estimate.

\citet{hebert} fuse sensory data from stereo cameras, monocular images
and tactile sensors for simultaneously estimating the pose of the
object and the robot arm with an Unscented Kalman Filter (UKF). Using all these
sensor modalities, the approach allows to track the object well
while it is being held by the robot. It is not clear how well it works when the object is not being held by the robot, since
in that case less information about the motion is available. Our
method uses the knowledge of the control inputs as well, when available, but it is able to track an object even when it is not being held by the robot.

\citet{ueda} uses a sampling-based approach. The object is represented as a partial point cloud 
and tracking is performed using a particle filter. The likelihood of an observation is computed
using a function of the squared distance between each point in the object model and its closest point in the 
observed point cloud, as well as the distance between their respective colors. This approach neither 
models noise in the sensor nor occlusion. Since in terms of
experimental results there is only a video available, 
it is not clear how well this method performs in general settings.

In contrast to the mentioned algorithms, occlusion is modeled explicitly in our observation model.
As we will show in Sect.~\ref{sec:obs} in more detail, this leads to strong
non-linearities. We therefore use a sampling-based approach to filter the pose of the object. 
This has the advantage that even non-Gaussian multi-modal distributions can be represented. The main drawback of sampling-based
approaches is that they are computationally expensive. Nevertheless we 
show in the experimental section that we achieve real time tracking using 
only one CPU core.

\section{Bayes Filter}\label{sec:bayes_filter}
In this section, we briefly discuss the Bayes filter and a number of techniques to perform inference in cases where the problem cannot be solved analytically.

The assumptions made in a Bayes filter are twofold. Firstly,
the Markov assumption asserts that each state only depends on the
previous state. Secondly, it is assumed that the state is sufficient to predict the (potentially noisy)
observation~\cite{thrun}.
\begin{figure}[h]
 \centering
 \includegraphics[width=0.4\textwidth]{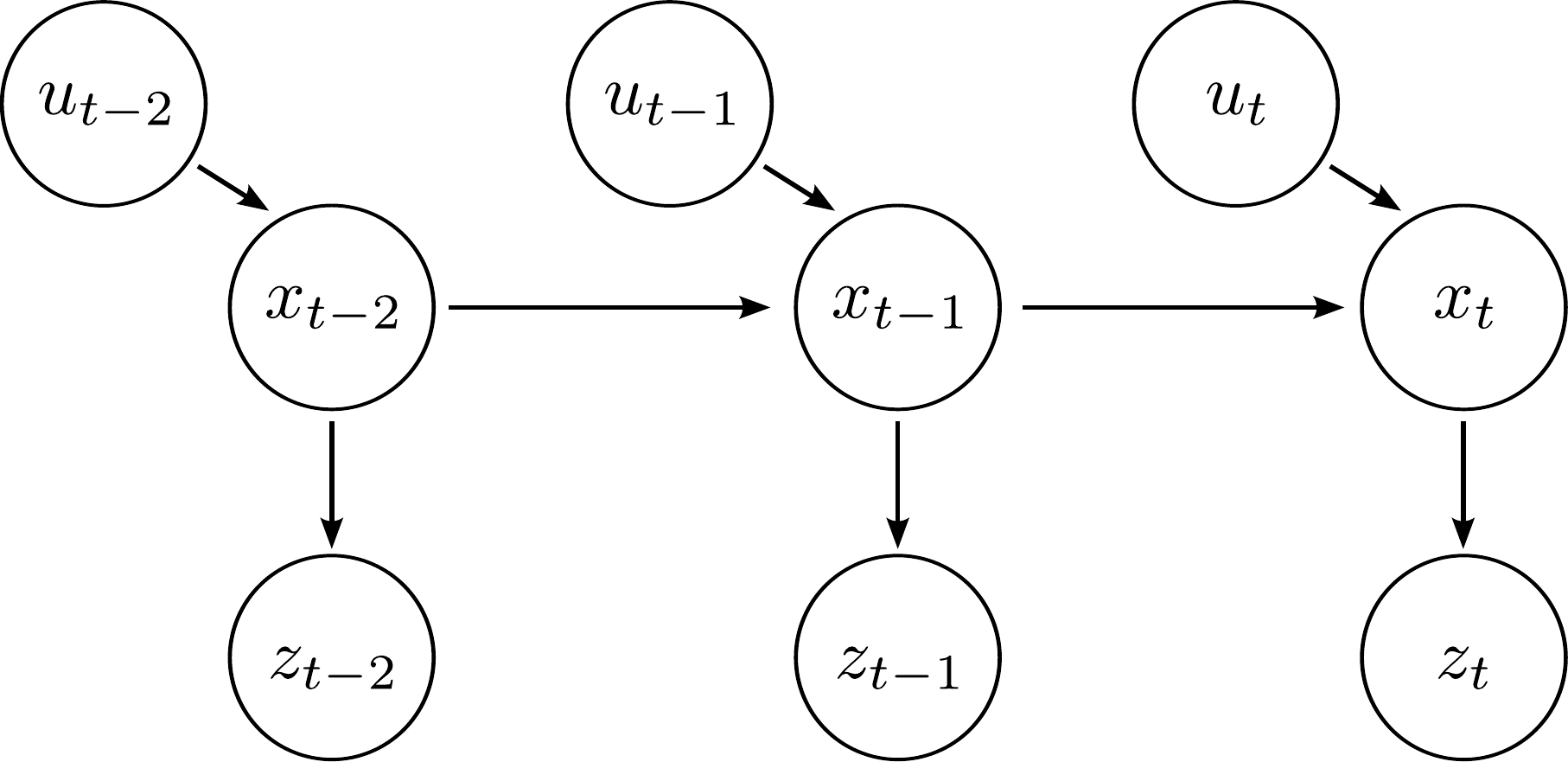}
 \caption{This graph represents the independence assumptions made by a Bayes filter.}
 \label{fig:graphical_model_bayes_filter}
\end{figure}
 These independence assumptions are represented in a probabilistic graphical
model (PGM) in Fig.~\ref{fig:graphical_model_bayes_filter}. The goal
in state estimation is to find the posterior
$p(x_{t}|z_{1:t},u_{1:t})$ over the state $x_{t}$
given all the observations $z_{1:t}$ and the control inputs
$u_{1:t}$. 
There are different approaches for
performing inference in this model that depend on the form of the process model $p(x_t|x_{t-1},u_t)$ and the observation model $p(z_t|x_t)$.
For a linear process and observation model and each being subject to
Gaussian noise, the inference
problem can be optimally solved with a Kalman Filter.

If the process and observation model can be approximated as being locally linear
and the noise can be assumed to be Gaussian, then an Extended Kalman
Filter (EKF) or an Unscented Kalman Filter (UKF) can be used. 
In the EKF, the process and observation model are
linearized to compute the covariance matrix of the current state
estimate. In the UKF, samples around the mean of the current state
estimate or prediction
are drawn and pushed through the non-linear models. The projected
samples serve as the base to estimate the Gaussian posterior distribution.

Finally, there are nonparametric methods to solve the inference problem. The
most well known is the particle filter which represents the posterior over the
state at time $t$,  $p(x_{ t }|z_{ 1:t },u_{ 1:t })$, by a set of samples $\{x_t^{(l)}\}$ which are distributed accordingly. Such a set of samples can be obtained by sampling from the distribution over the entire sequence of states $p(x_{ 1:t }|z_{ 1:t },u_{ 1:t })$ and then dropping all the previous states $x_{ 1:t-1 }$. Consequently, the previous states do not have to be marginalized out in a particle filter,  we can thus express $p(x_{ 1:t }|z_{ 1:t },u_{ 1:t })$ instead of $p(x_{t}|z_{1:t},u_{1:t})$ \cite{thrun}.

As will be discussed in more detail in Sect.~\ref{sec:obs}, we introduce a variable $o_t$ describing which parts of the image are occluded, therefore our state $x_t$ consists of the 6-DoF pose and the occlusion $(r_t, o_t)$.
The posterior distribution $p(x_{ 1:t }|z_{ 1:t },u_{ 1:t })$ can then be
written as
\begin{align}
 p(r_{1:t}, o_{1:t}|z_{1:t},u_{1:t})=p(o_{1:t}|r_{1:t},z_{1:t},u_{1:t})p(r_{1:t}|z_{1:t},u_{1:t})\notag.
\end{align}
As a result of the functional form of our observation model, derived in Sect.~\ref{sec:obs}, and our process model, discussed in Sect.~\ref{sec:process}, the variables $o_{1:t}$ can be marginalized out
analytically while the variables $r_{1:t}$ cannot. In a
dynamic Bayesian network with these properties, inference can be
performed using a Rao-Blackwellised particle filter \cite{doucet}. Integrating out the previous occlusions $o_{1:t-1}$ we obtain
\begin{align}
 \!\!\!p(r_{1:t}, o_{t}|z_{1:t},u_{1:t})\!=\!p(o_{t}|r_{1:t},z_{1:t},u_{1:t})p(r_{1:t}|z_{1:t},u_{1:t})\label{eq:rao}
\end{align}
Since the variables $r_{1:t-1}$ cannot be integrated out analytically, the second term is represented, as in a common particle filter, with a set of samples. In a Rao-Blackwellised particle filter the posterior over the full state $p(r_{t}, o_{t}|z_{1:t},u_{1:t})$  is thus represented by a set of particles $r_{1:t}^{(l)}$ distributed according to $p(r_{1:t}|z_{1:t},u_{1:t})$, each associated with a  probability $p(o_{t}|r_{1:t}^{(l)},z_{1:t},u_{1:t})$.


Before we apply this method to object tracking we will derive the observation and the process model.

\section{OBSERVATION MODEL}\label{sec:obs}
\begin{figure}[h]
 \centering
 \includegraphics[width=0.4\textwidth]{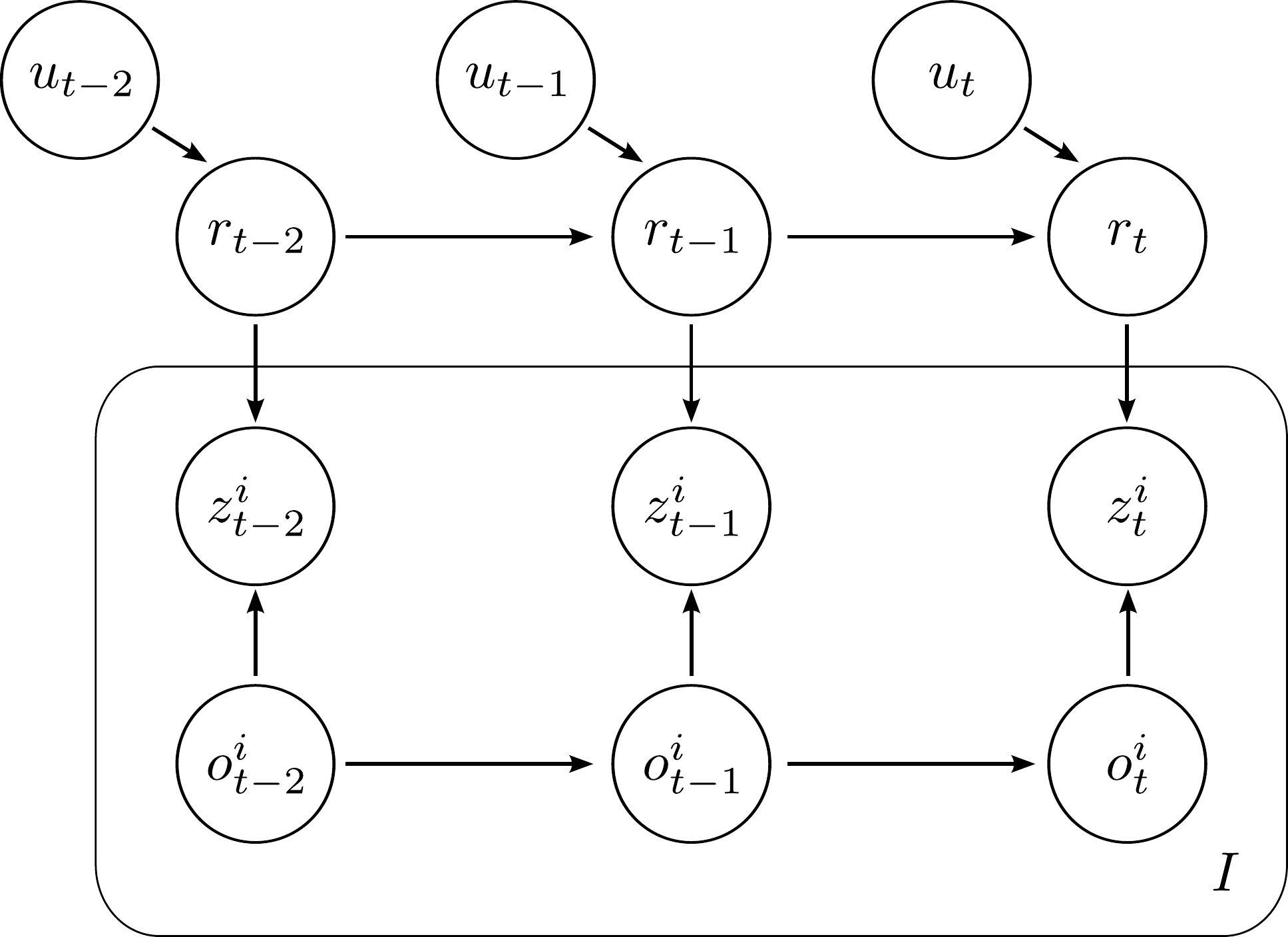}
 \caption{This graph represents the independence assumptions made by a our filter. The box is a plate which represents the $I$ pixels.}
 \label{fig:graphical_model}
\end{figure}

The objective is to infer the 6-DoF pose $r_t$ of an object assuming
that we have a 3D model. The observation model $p(z_t|r_t)$ expresses
what depth image we expect to observe given the pose of the
object. However, if it is not known whether the object is occluded or
not, we cannot predict the depth which should be measured. Therefore,
we introduce a set of binary variables $o_t=\{o_t^i\}$ modelling
occlusion. $o_t^i=0$ means that the object is visible at time $t$ in
pixel $i$, whereas $o_t^i=1$ indicates that it is occluded. Thus, the
full state $x_t=(r_t,o_t)$ consists of the 6-DoF pose and the
occlusion for each pixel $i$. The graphical model in
Fig.~\ref{fig:graphical_model_bayes_filter} can be expanded as shown in Fig.~\ref{fig:graphical_model}.
The depth measurements $\{z_t^i\}$ at each pixel (including the noise) are assumed to be independent given
the state. This assumption is reasonable since the pose and the occlusion variables allow us to predict
the observation. 
Furthermore, we assume the occlusions ${o_t^i}$ of
different pixels to be independent. 
Although we thereby ignore relations between neighboring pixels, inference
becomes more efficient as will be described in Sect.~\ref{sec:filter}.

We can now write  $p(z_t|r_t,o_t)=\prod_i p(z_t^i|r_t,o_t^i)$ as the
product of measurement likelihoods at each pixel $i$ given the pose
and whether the object is occluded in pixel $i$. By marginalizing out
the occlusion variables $p(z_t^i|r_t) =
\sum_{o_t^i}p(z_t^i|r_t,o_t^i)p(o_t^i)$, we obtain a model which is
closely related to beam-based models discussed in robotics literature
\cite{thrun}. These consist of a weighted sum of an observation model
assuming that the object is occluded and an observation model assuming
that it is visible. The important difference to our approach is that
we continuously estimate the probability $p(o_t^i)$ of the object being occluded whereas in \cite{thrun} this is a parameter which is set off-line and kept constant during execution.

\begin{figure}[h]
 \centering
 \includegraphics[width=0.4\textwidth]{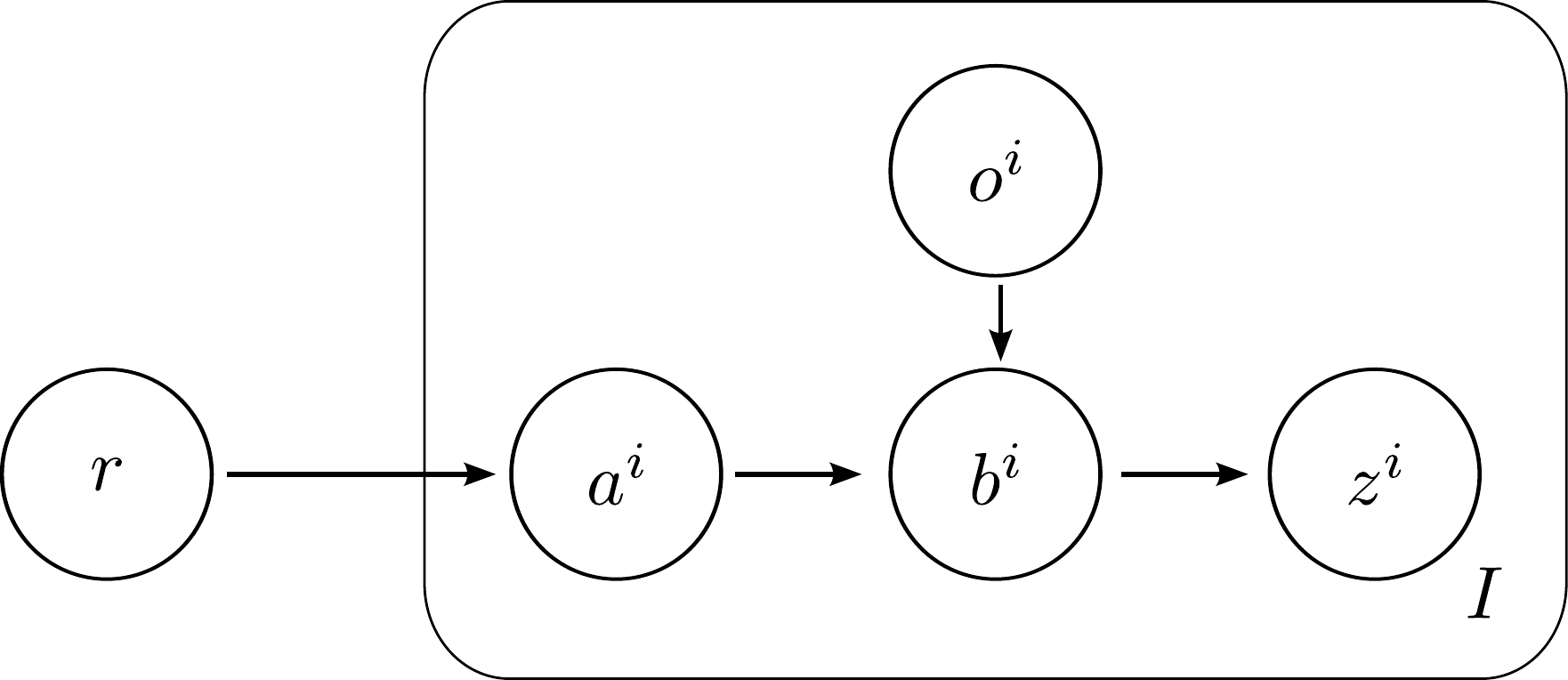}
 \caption{This graph shows the measurement process of the camera. It is a subpart of the graph in Fig.~\ref{fig:graphical_model}, zoomed in on one observation.}
 \label{fig:graphical_model_observation}
\end{figure}

To express the  observation model $p(z_t^i|r_t,o_t^i)$, we represent the measurement process in Fig.~\ref{fig:graphical_model_observation}, which is a subpart of the graphical model in
Fig.~\ref{fig:graphical_model}.
 Two auxiliary variables $a^i$ and $b^i$ have
been introduced. These variables will be integrated out, but they are convenient for
the formulation of the observation model. Since we are looking at only one time step, the time index is omitted here. $a^i$ is the distance to the tracked object along the beam defined by pixel $i$ and $b_i$ is the distance to the object which is seen in pixel $i$; this might not be the tracked object, depending on whether it is occluded.

We can now describe the process which leads to the observation. Given
the pose $r$, we can compute the distance to the tracked object $a^i$
for a given pixel $i$. Then given the distance to the tracked object
and the information whether it is occluded, we can predict the
distance to the object $b^i$ seen in pixel $i$, which might or might
not be the tracked object. Knowing this distance, it is easy to
predict a measurement $z^i$. Using the independence assumptions from
Fig.~\ref{fig:graphical_model_observation}, this process can be
formally written as
\begin{align}
 p(z^{ i }|r,o^{ i })&=\int _{ a,b } p(z^{ i }|b^{ i })p(b^{ i }|a^{ i },o^{ i })p(a^{ i }|r).
\end{align}
In the following, we express each of these terms going from right to
left. $p(a^{ i }|r)$ is the probability distribution over the distance
to the tracked object given the pose. This is simply the distance $d^i(r)$
to the intersection of the beam defined by pixel $i$ with the object
model in pose $r$. Additionally, there is some noise due to errors in the object model:
\begin{align}
 p(a^{ i }|r) = \mathcal{N}(a^i|d^i(r), \sigma_m).
\end{align}
$p(b^{ i }|a^{ i },o^{ i })$ expresses the distance to the object measured in pixel $i$ given the distance to the tracked object and the occlusion. If the $o^{ i }=0$ the object is visible, therefore $a^i$ and $b^i$ have to be identical. Otherwise $b_i$ has to be smaller than $a_i$, since the occluding object is necessarily in front of the tracked object. Formally we can write this as 
\begin{align}
 p(b^{ i }|a^{ i },o^{ i }) = 
  \begin{cases}
   \delta(b^{ i }-a^{ i }) & \!\! \text{if } o^{ i }\! =\! 0 \\
   I(b^{ i }>0 \land b^{ i }<a^{ i })\frac { \lambda e^{ -\lambda b^{ i } } }{ 1-e^{ -\lambda a^{ i } } }       & \!\! \text{if } o^{ i } = 1
  \end{cases}
    \label{eq:distance_to_first_object}
\end{align}
where $I$ is the indicator function which is equal to one if the condition is true and zero otherwise. $\delta$ denotes the Dirac delta function. We assume that the probability of a beam intersecting with an object other than the tracked object is equal for any interval of equal length. This implies that the probability of the beam hitting the first object at a distance $b^{ i }$ decays exponentially if the observed object is not the tracked object, see Eq.~\ref{eq:distance_to_first_object}. The parameter $\lambda$ is a function of the half-life, i.e. the distance at which we expect half of the beams to not have intersected with an object yet. The algorithm is not sensitive to this parameter, therefore we simply set the half-life to a number which seems reasonable. In all our experiments we used \unit[1]{m}.

Finally, we express $ p(z^{ i }|b^{ i })$ which is the distribution
over the measurement $z^{ i }$ given the distance of the observed
object to the camera. The difference in these two quantities is due to
noise in the measurement of the range camera. We used an Asus XTION
Pro depth sensor in our experiments and thus attempted to model its
noise. Khoshelham and Elberink~\cite{kourosh12} estimated the noise in
the depth measurements of a Microsoft Kinect which is based on the
same design as the sensor we used. The authors showed that the noise
increases with the depth squared. We define the camera standard
deviation $\sigma_c$ accordingly. Fallon et al.~\cite{fallon} model
the noise in the Kinect camera by the sum of a Gaussian distribution
and a constant term to shift more weight to the tails than in a purely
normal distribution. We also found it to be advantageous to have a
heavy tailed distribution and thus define the distribution in a
similar fashion as
\begin{align}
 p(z^{ i }|b^{ i })&\!\!=\!\!(1-\beta )\mathcal{ N }(z^{ i }|b^{ i },\sigma _{ c })+\beta \frac { I(z^{ i }>0\land z^{ i }<m) }{ m } 
\end{align}
where $I$ is the indicator function and $m$ is the maximum depth which can be measured by the range camera, \unit[6]{m} in our case. In all our experiments the weight of the tails $\beta$ was set to $0.01$, as proposed in \cite{fallon} and $\sigma_c$ was defined according to \cite{kourosh12}.

\begin{figure}[h]
 \centering
 \includegraphics[width=0.48\textwidth]{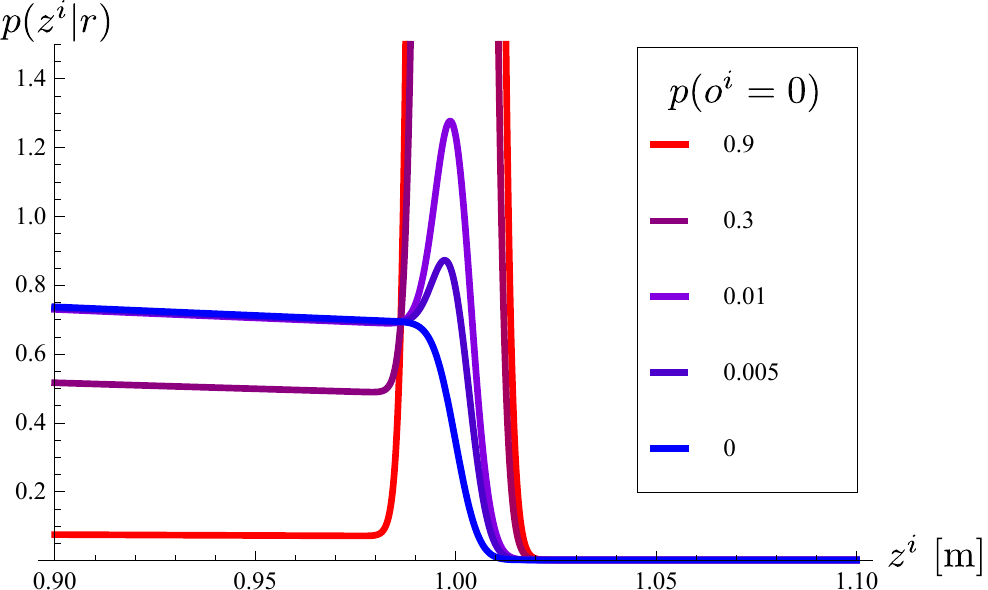}
 \caption{This graph shows the probability of an observation given a pose $p(z^i|r)=\sum_{o^i}p(z^i|r,o^i) p(o^i)$. The pose $r$ is such that the distance to the object model $d^i(r)$ is equal to \unit[1]{m}. 
 There are five curves for different probabilities of the object being visible $p(o^i=0)$.}
 \label{fig:likelihood_occlusion}
\end{figure}

Now the observation model is fully specified. We plot the likelihood of the pose $p(z^i|r)=\sum_{o^i}p(z^i|r,o^i) p(o^i)$ in Fig.~\ref{fig:likelihood_occlusion} as a function of the depth measured in pixel $i$, given that the predicted distance to the object $d^i(r)$ is equal to \unit[1]{m}.
  For a small probability of the object being visible the function is almost constant in observed depth $z^i$ with a steep decrease at the predicted depth $d^i(r)=1$. This intuitively makes sense since the occluding object has to be closer to the camera than the tracked object. As the probability of the object being visible increases, a more and more pronounced spike appears at the predicted depth. At the same time the probability of measuring a depth which is smaller than the predicted depth diminishes as we become more certain that the object is not occluded.

\section{PROCESS MODELS}\label{sec:process}
The pose process model $p(r_t|r_{t-1},u_t)$ describes how we expect
the object pose to change over time given the last pose and the
control inputs. The occlusion process model $p(o_t^i|o_{t-1}^i)$
describes how the occlusions propagate. In this section we will model these two probability distributions.\\
\subsection{Pose Process Model}
The process model for the pose $p(r_t|r_{t-1},u_t)$ depends of course
on the underlying process. In our experiments we test two different
scenarios. In the first case where the object is moved by a human arm, we do thus not have access to the control inputs and therefore simply propagate the pose by a small random rotation and translation drawn from a Gaussian distribution. 

In the second case, the object is moved by a robot and we thus make use of the control inputs. From the control inputs, we can compute the expected velocity of the object being held for every time step. Since this velocity is not entirely accurate, we model the noise as being Gaussian in the translational as well as the rotational velocity. When the object is being moved by a robot, the uncertainty is of course much smaller than when it is being moved by a human arm.
\subsection{Occlusion Process Model}
The distribution $p(o_t^i|o_{t-1}^i)$ has only two parameters since
the variables are binary. We have to determine the probability of the
object being visible in pixel $i$ given that it was visible in the
last time step, $p(o_t^i=0|o_{t-1}^i=0)$ and the probability that the
object is visible given that is was occluded in the last time step
$p(o_t^i=0|o_{t-1}^i=1)$. In all our experiments we set
$p(o_t^i=0|o_{t-1}^i=0) = 0.9$ and $p(o_t^i=0|o_{t-1}^i=1) = 0.3$ for
a time delta of one second. This represents the fact that if the
object was observed in the last time step we expect to observe it
again in the current time step. The algorithm proved to not be very
sensitive to these parameters, so their values were simply set by hand
without any tuning. A more systematic estimation is the subject of future work.

\section{FILTERING ALGORITHM}\label{sec:filter}
Since we have introduced a new set of variables, we do not only infer the pose of the object $r_t$ but also the occlusion $o_t$.
As can be seen from Fig.~\ref{fig:likelihood_occlusion}, our observation model is
highly nonlinear and results in a likelihood of the pose $p(z_t|r_t)$ which
does not fit well to a normal distribution. With a Gaussian
distribution, we can essentially express that a variable should be somewhere close to a
certain value. However as discussed in Sect.~\ref{sec:obs}, for the
proposed algorithm it is central to also be able to
express that a variable should be larger than a certain value. This requirement discards algorithms which
approximate the $p(z_t|r_t)$ with a normal distribution such as KF, EKF and UKF.
Therefore, we use sampling to solve the inference problem for the pose $r_t$. As shown below, marginalizing out the occlusion variables $o_t$ is tractable. We can thus perform inference using a Rao-Blackwellised particle filter as described in Sect.~\ref{sec:bayes_filter}.
Rewriting Eq.~\ref{eq:rao} for convenience, we have
\begin{align}
\!\!p(r_{1: t },o_{ t }|z_{ 1:t },u_{ 1:t })\!\!= \!p(o_{ t }|r_{ 1:t },z_{ 1:t },u_{ 1:t })p(r_{ 1:t }|z_{ 1:t },u_{ 1:t })\label{eq:joint_prob}
\end{align}
We are mainly interested in the second term which expresses the
distribution over the poses given all the measurements and control
inputs. As we will see shortly, the estimate of the pose depends on
the estimate of the occlusion i.e. the first term in
Eq.~\ref{eq:joint_prob}. In the following, we will therefore express these two terms recursively.

From Fig.~\ref{fig:graphical_model} we can see that the first term in Eq.~\ref{eq:joint_prob} factorizes as $p(o_{ t }|r_{ t },z_{ 1:t },u_{ 1:t }) = \prod_i p(o_{ t }^{ i }|r_{ 1:t },z_{ 1:t },u_{ 1:t })$. Taking the independence assumptions from Fig.~\ref{fig:graphical_model} into account, we can express each factor in a recursive form.
\begin{align}
 &p(o_{ t }^{ i }|r_{ 1:t },z_{ 1:t },u_{ 1:t })=\label{eq:observability_propagation}\\
 &\frac { \sum _{ o_{ t-1 }^{ i } } \!\left[ p(z_{ t }^{ i }|r_{ t },o_{ t }^{ i })p(o_{ t }^{ i }|o_{ t-1 }^{ i })p(o_{ t-1 }^{ i }|r_{ 1:t-1 },z_{ 1:t-1 },u_{ 1:t-1 })\! \right]}{ \sum _{ o_{ t }^{ i },o_{ t-1 }^{ i } } \!\!\! \left[ p(z_{ t }^{ i }|r_{ t },o_{ t }^{ i })p(o_{ t }^{ i }|o_{ t-1 }^{ i })p(o_{ t-1 }^{ i }|r_{ 1:t-1 },z_{ 1:t-1 },u_{ 1:t-1 })\! \right]} \notag
\end{align}
 where $p(z_{ t }^{ i }|r_{ t },o_{ t }^{ i })$ represents the observation model (see Sect.~\ref{sec:obs}), and $p(o_{ t }^{ i }|o_{ t-1 }^{ i })$ is the occlusion process model (see Sect.~\ref{sec:process}).
 
Taking the independence assumptions from Fig.~\ref{fig:graphical_model} into account, we can write the second term from Eq.~\ref{eq:joint_prob} as
\begin{align}
&p(r_{ 1:t }|z_{ 1:t },u_{ 1:t })\propto \label{eq:filter} \\
&p(z_{ t }|r_{ 1:t },z_{ 1:t-1 })p(r_{ t }|r_{ t-1 },u_{ t })p(r_{ 1:t-1 }|z_{ 1:t-1 },u_{ 1:t-1 }) \notag
\end{align}
where $p(r_{ t }|r_{ t-1 },u_{ t })$ is the pose process model as discussed in Sect.~\ref{sec:process}.
We can obtain a set of samples distributed
according to $p(r_{ 1:t }|z_{ 1:t },u_{ 1:t })$ by taking the samples
from the previous time step, propagating them with the process model
$p(r_{ t }|r_{ t-1 },u_{ t })$ and resampling using the likelihoods
$p(z_{ t }|r_{ 1:t },z_{ 1:t-1 })$. In a common particle filter the likelihood
$p(z_{ t }|r_{ 1:t },z_{ 1:t-1 })$ reduces to $p(z_{ t }|r_{t})$, but here it depends on the estimate of the occlusion:
\begin{align}
 &p(z_{ t }|r_{ 1:t },z_{ 1:t-1 })=\\
 &\sum _{ o_{ t-1 },o_{ t } } p(z_{ t }|r_{ t },o_{ t })p(o_{ t }|o_{ t-1 })p(o_{ t-1 }|r_{ 1:t-1 },z_{ 1:t-1 },u_{ 1:t-1 }). \notag
\end{align}
The sums which marginalize out the occlusion variables contain $2^I$
terms, with $I$ being the number of pixels. This is of course
intractable, but given our assumptions, all the terms inside of the
sum factorize over the pixels $i$, and we can move the sum inside of
the product.
\begin{align}
 &p(z_{ t }|r_{ 1:t },z_{ 1:t-1 })= \label{eq:likelihood}\\
 &\prod _{ i } \!\! \sum _{ o_{ t }^{ i },o_{ t-1 }^{ i } } \!\!\!\!\! \left[ p(z_{ t }^{ i }|r_{ t },o_{ t }^{ i })p(o_{ t }^{ i }|o_{ t-1 }^{ i })p(o_{ t-1 }^{ i }|r_{ 1:t-1 },z_{ 1:t-1 },u_{ 1:t-1 }) \right] \notag
\end{align}
Now we only have to sum $I$ times over $4$ terms. The last term is identical to Eq.~\ref{eq:observability_propagation}, just shifted by one time step. The estimate of the pose thus depends on the estimate of the occlusion through the likelihood.\\

The initialization of the tracking is not a focus of this paper. In practice we tackle this problem by putting the object in a configuration where we have a strong prior over its pose, for example when it is standing on a table or being held by the robot. We then initialize the filter with a very large number of particles sampled from this prior. The initialization could also come from a different algorithm which is better suited for a global search.

Now the algorithm is fully defined:
\begin{itemize}
 \item From the previous time step we have a set of particles $\{r^{(l)}_{1:t-1}\}$ distributed according to $p(r_{ 1:t-1 }|z_{ 1:t-1 },u_{ 1:t-1 })$ and for each of these particles we know $p(o_{ t-1 }^{ i }|r_{ 1:t-1 }^{(l)},z_{ 1:t-1 },u_{ 1:t-1 })$. Furthermore we know the control $u_t$ which is applied during the current time step and we observe a depth image $z_t$.
 \item For each particle in $\{r^{(l)}_{1:t-1}\}$
 \begin{itemize}
  \item We draw a sample $r_t^{(l)}$ from the pose process model $p(r_t|r_{t-1}^{(l)},u_t)$.
  \item We compute the likelihood $p(z_{ t }|r_{ 1:t }^{(l)},z_{ 1:t-1 })$ according to Eq.~\ref{eq:likelihood}.
  \item We update the occlusion probabilities $p(o_{ t }^{ i }|r_{ 1:t }^{(l)},z_{ 1:t },u_{ 1:t })$ for each pixel according to Eq.~\ref{eq:observability_propagation}.
 \end{itemize}
 \item We resample the particles according to the likelihoods. We thus now have a set of particles $\{r^{(l)}_{1:t}\}$ distributed according to $p(r_{ 1:t}|z_{ 1:t },u_{ 1:t })$, and the corresponding occlusion probabilities $p(o_{ t }^{ i }|r_{ 1:t }^{(l)},z_{ 1:t },u_{ 1:t })$. We can now go to the next time-step and repeat the procedure above.
\end{itemize}

\section{EXPERIMENTAL RESULTS}\label{sec:exp}

In the following, we describe our experimental setup and present results. Due to
the difficulty in obtaining ground truth information, we use a previously
implemented fiducial-based object tracker as a baseline method. We want to
emphasize that our approach does not rely on the presence of these fiducials.

\subsection{Experimental Setup}

Our experiments are based on the dual arm manipulation platform shown in
Fig.~\ref{fig:title}. The head is actuated by two stacked pan-tilt units, and
consists of the following sensors: (a) an Asus Xtion PRO range camera, (b) a
Point Grey Bumblebee2 stereo camera, and (c) a Prosilica high resolution color
camera.
These cameras are calibrated w.r.t each other using an offline calibration
procedure. The robot has two 7-DoF Barrett WAM arms, each equipped with a
3-fingered Barrett Hand as the end-effector. The WAM arms are cable driven, and
joint positions are measured using absolute encoders on the motors and not the
joints themselves. The variable stretch of the cables depending on the robot
pose and payload imply that the kinematic model of the arm is not accurate
enough to perform manipulation tasks with sufficient precision~\cite{Pastor13}.

We use the Asus Xtion as the sensor for our algorithm, which provides depth
images at a rate of \unit[30]{Hz}. The depth images produced by this camera
have a resolution of 640x480, for our purposes however an image which has been 
downsampled by a factor of 5 proved to be detailed enough. 
Our observation is thus an array of size 128x96 containing the depth
measurements from the camera. We would like to stress that apart from the downsampling
no preprocessing whatsoever is required by the proposed method.
Our implementation currently uses a single
core of a 3.3GHz Intel Xeon W5590 CPU, and is able to sample and evaluate 200
poses in real-time per camera frame.
\begin{figure}[h]
 \vspace{0.2cm}
 \centering
 \includegraphics[width=\linewidth]{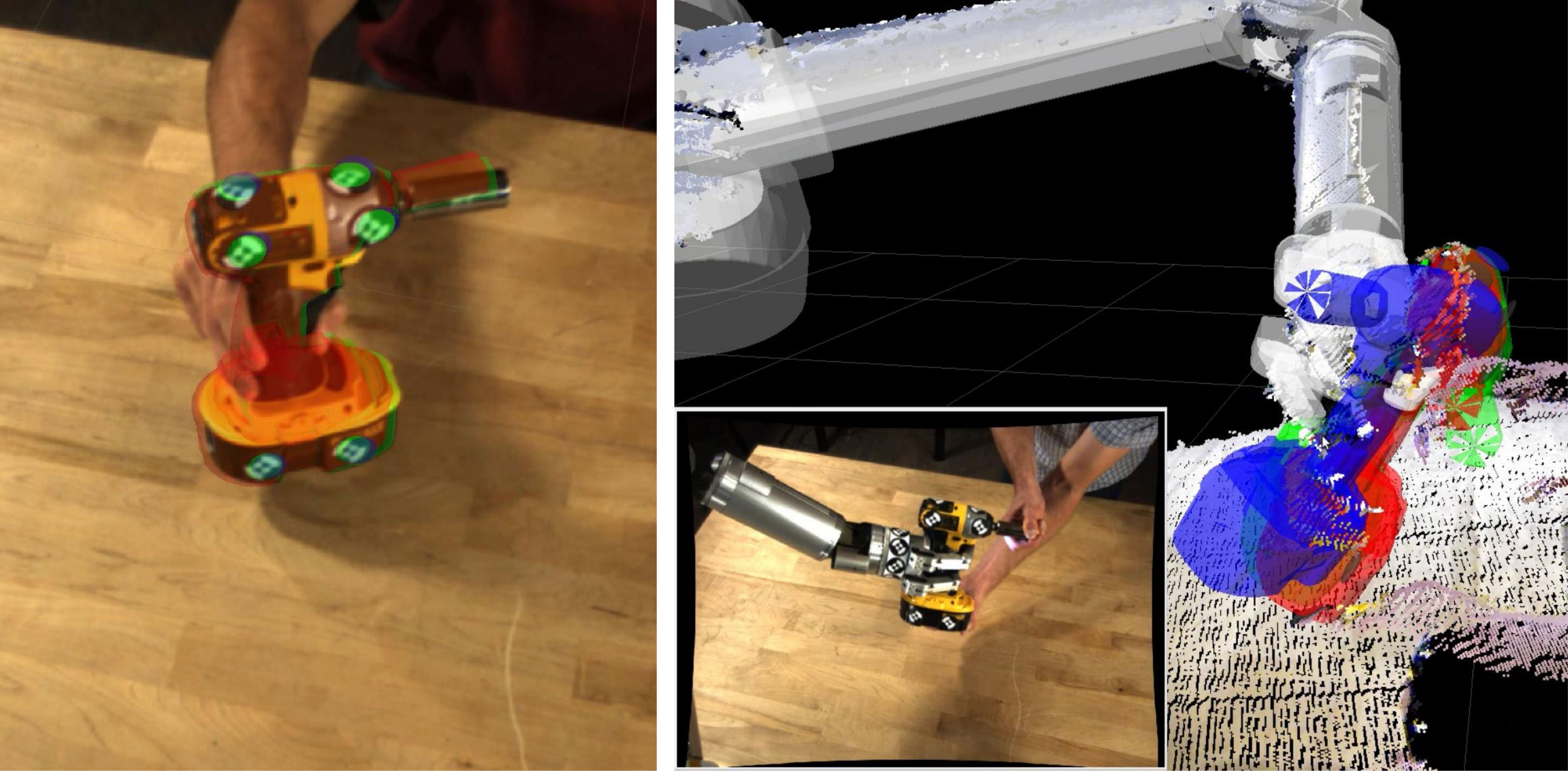}
 \vspace{-0.7cm}
 \caption{The fiducial tracker uses camera images to track the round fiducial
 markers (left). The obtained pose estimate (green) is used as a baseline comparison. The forward kinematics model (blue) is misplaced because of deliberate disturbances (small image). The proposed approach achieves accurate tracking (red) even in the presence of occlusions.}
 \label{fig:experiment}
\end{figure}
We compare the results from our method with a
previously implemented baseline method which tracks objects based on a known
pattern of fiducials on the object, using the Bumblebee2 stereo camera. This
method is based on an EKF which maintains a distribution over the pose of the
object. Fiducials are detected in each camera image using a local
template-matching approach\footnote{The fiducial detector was kindly provided by Paul Hebert, Jeremy Ma, and Nicolas Hudson from the Jet Propulsion Laboratory.} (see Fig.~\ref{fig:experiment} left).
\begin{figure}[b!]
 \centering \includegraphics[width=0.99\linewidth]{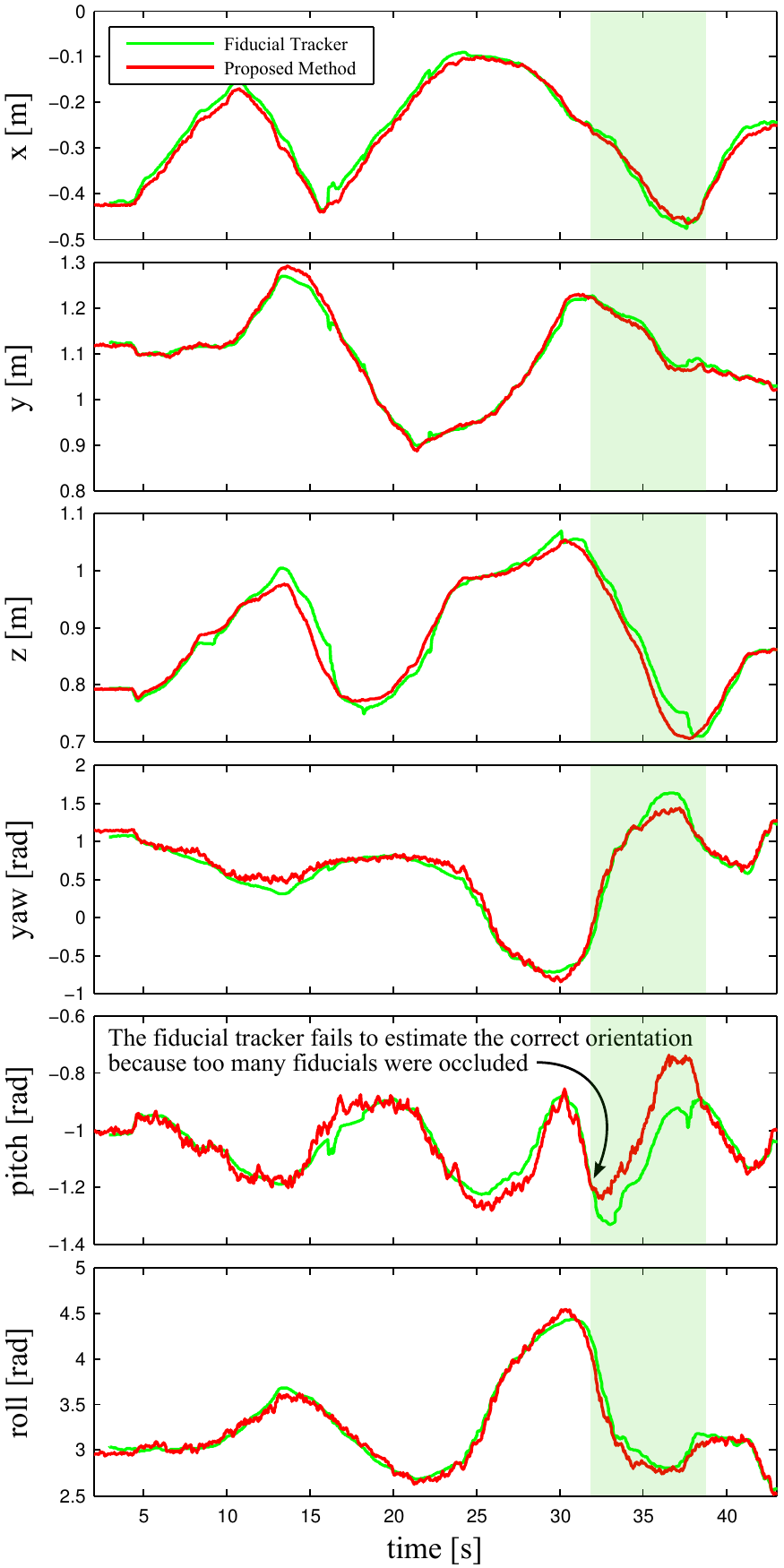}
 \vspace{-0.6cm}
 \caption{Scenario (a): 6-DoF trajectories of the object being manipulated by a
 human hand, as tracked by the proposed method (red) and the baseline fiducial
 tracking method (green). The fiducial tracking method is susceptible to errors
 when fiducials are occluded (e.g., from $t$=33 to $t$=37), while our method is
 more robust to occlusions.}
 \label{fig:human_hand}
\end{figure}
The template for each fiducial depends on the projection of the expected fiducial pose in the
camera. The 3-D position of each fiducial is then reconstructed from their 2-D
positions in each camera. The full 6-DoF object pose is solved for by minimizing
the squared error between the detected fiducial positions and their
corresponding positions in the object model. The object pose thus obtained is
treated as an observation which updates the pose distribution in the EKF. This
EKF uses the same process models as the ones described in Sect.~\ref{sec:process},
depending on whether the object is in the robot hand or not. Finally, we also contrast these visual tracking methods with estimates of the object pose
based on the kinematic model of the robot. The position of the hand is computed
using the motor encoders. We compute a fixed offset between the hand
and the object at the time of grasping, and assume that this offset remains
constant for the entire motion.

\subsection{Results}

Our algorithm was evaluated in three scenarios: (a) object manipulated by a
human (Fig.~\ref{fig:experiment} left), (b) object manipulated by the robot with a secure grasp, and (c) object manipulated by the robot (Fig.~\ref{fig:experiment} right), with a human disturbing the pose of the object in the
hand. Videos of each of these scenarios may be found in the multimedia
attachment or online at~\mbox{\url{http://youtu.be/MBgggaJq1sY}}.

\begin{figure}[t]
 \centering
 \includegraphics[width=0.99\linewidth]{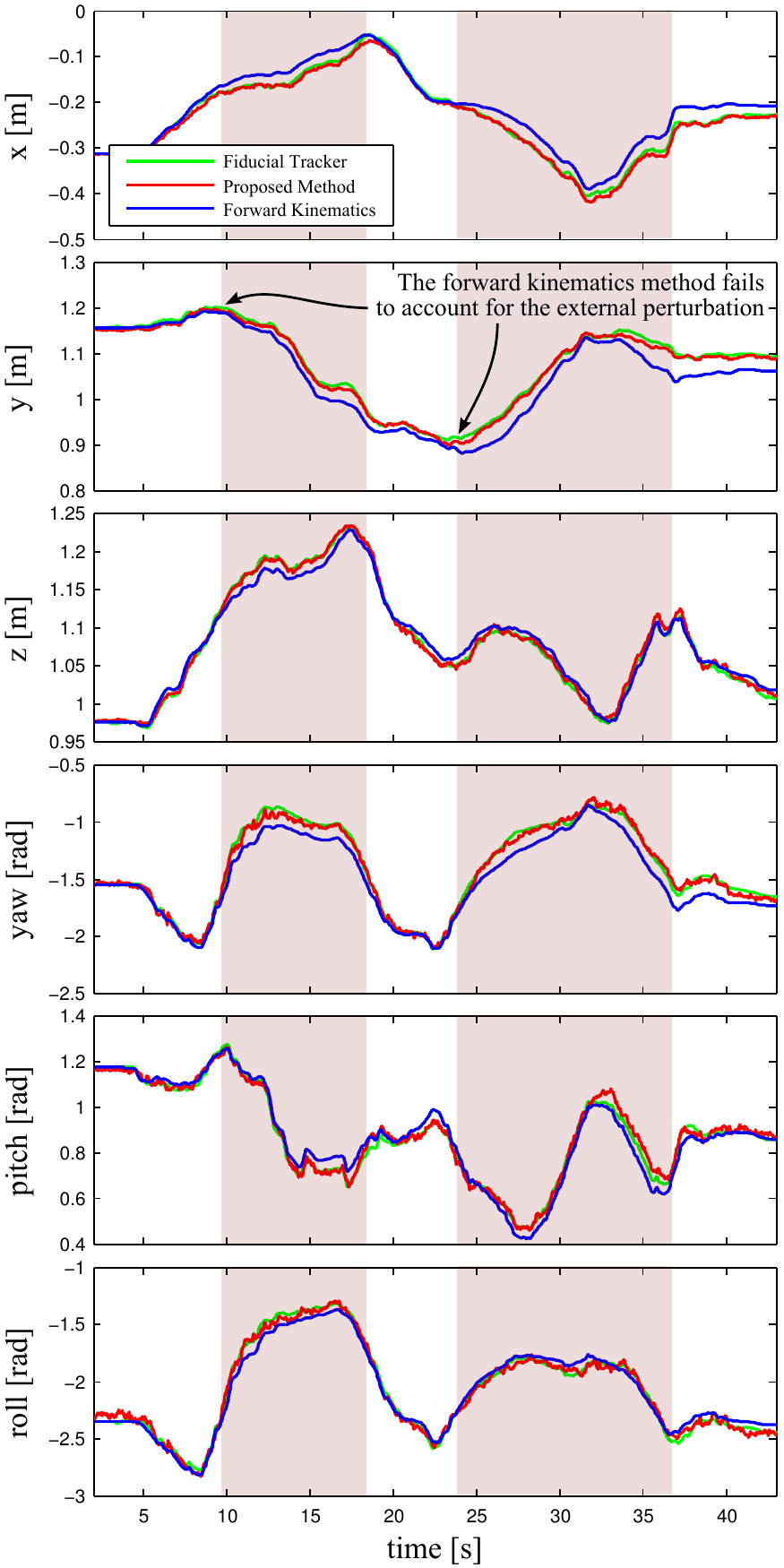}
 \vspace{-0.6cm}
 \caption{Scenario (b): 6-DoF trajectories of the object being manipulated by
 the robot with a secure grasp. The errors in the forward kinematics method
 (blue) are due to imprecise joint position sensing under load. The proposed
 approach (red) followed the baseline (green) closely.}
 \label{fig:robot_hand_no_slippage}
\end{figure}
\begin{figure}[t]
 \centering
 \includegraphics[width=0.99\linewidth]{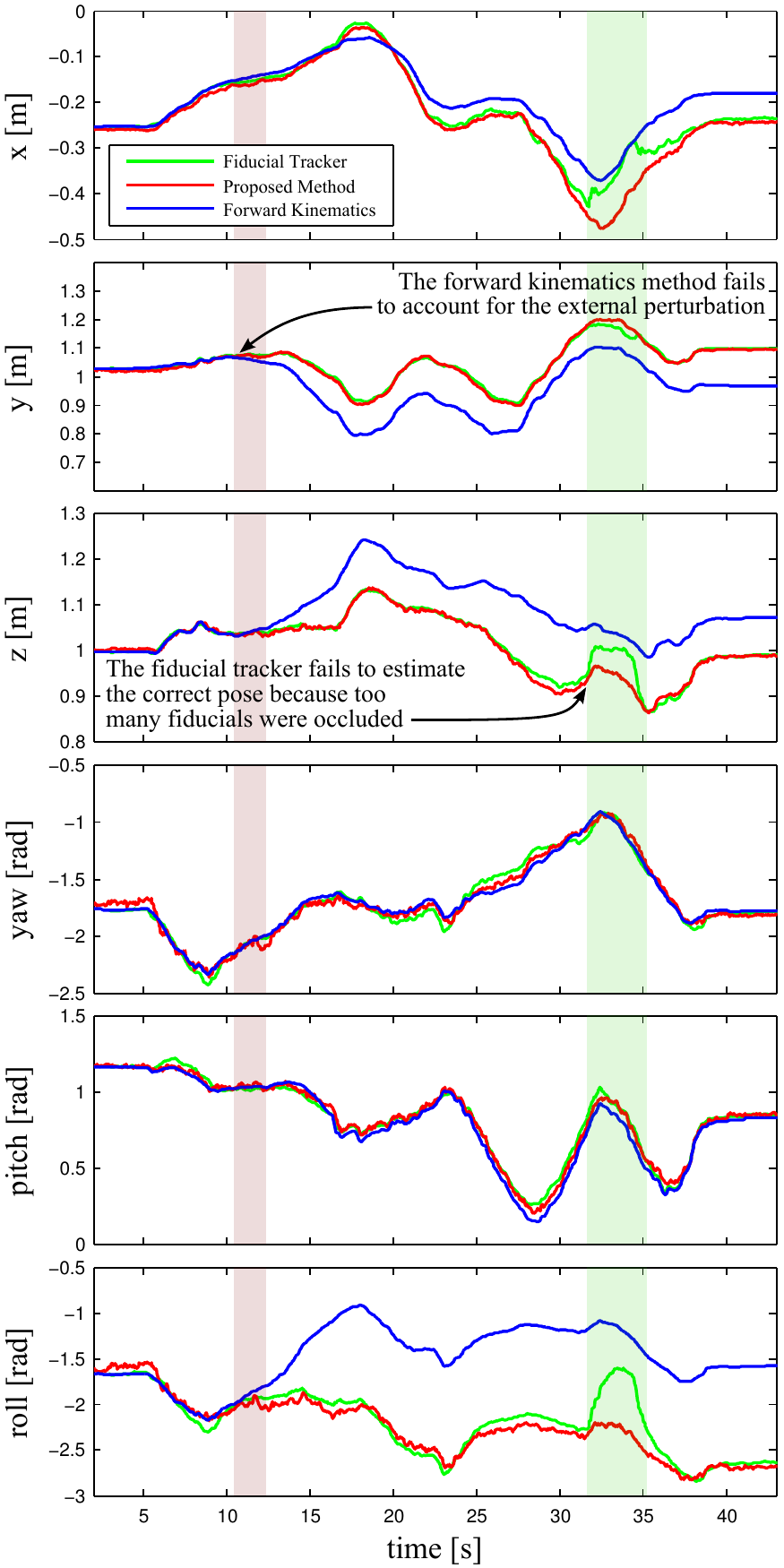}
 \vspace{-0.6cm}
 \caption{Scenario (c): 6-DoF trajectories of the object being manipulated by
 the robot, while being disturbed by a human. Changes in the pose of the object
 in the hand cannot be tracked by the forward kinematics method (blue). Our
 marker-less approach (red) successfully tracked the object and followed the
 baseline fiducial-based approach (green) closely.}
 \label{fig:robot_hand_slippage}
\end{figure}

Figure~\ref{fig:human_hand} shows the position (x, y, z) and orientation (roll, pitch, yaw) trajectories of the object being
moved by a human, as tracked by the baseline fiducial tracking method (shown in
green) and our proposed method (shown in red). A qualitative assessment reveals
that the fiducial tracking method is susceptible to errors when fiducials
are occluded (e.g., from $t$=33 to $t$=37), while the proposed method tracks the object pose more robustly in spite of occlusions.

Results from scenario (b) where the object is manipulated by the robot with a
secure grasp are shown in Figure~\ref{fig:robot_hand_no_slippage}. In addition
to the baseline method and the proposed method, we also show pose estimates of
the object based on the robot kinematic model (in blue). This estimate
significantly deviates from the two visual tracking methods due to undetected
stretching of the Barrett WAM cables, which is exacerbated by the weight of the
object in the hand. Additionally, we introduced external disturbances that did
not affect the pose of the object with respect to the hand. Such errors in the
forward kinematics model render precise manipulation and tool use impossible
without visual tracking and control.

\balance
Figure~\ref{fig:robot_hand_slippage} shows results from the final scenario, in
which the object is manipulated by the robot, while the pose of the object in
the hand is being disturbed by a human. By definition, manipulation tasks
involve contact of the object in the hand with the external world. These
contacts often result in slippage of the object in the hand, and in such cases
visual tracking can be tremendously beneficial for successful completion of the
task. Similar to the previous scenario, we notice that the two visual tracking
methods perform equally well. In contrast, the forward kinematics method shows
large errors as it cannot account for movement of the object in the hand (see Fig.~\ref{fig:experiment} right).

\section{CONCLUSION AND FUTURE WORK}
\label{sec:conclusion}

We have presented a probabilistic approach to object tracking using a range
camera. Occlusions are explicitly modeled in our approach, which adds robustness
and removes the need to filter out points belonging to the robot. Our method is
fast enough for real-time tracking performance on a single core of a modern
computer, and the sampling process trivially lends itself to parallelization.

In future work, we plan to apply this method to tracking the posture of
articulated objects such as the robot arm itself to accurately estimate the
arm kinematic configuration. 
We also plan to apply the probabilistic models developed in this work to
alternate sensor modalities such as tactile sensors.
Finally, we will use the presented object tracking approach to further close perception-action loops as in~\cite{Pastor_2011}.


\section*{ACKNOWLEDGMENTS}
{\linespread{1.0} \small
We would like to thank Paul Hebert, Jeremy Ma, and Nicolas Hudson from the Jet
Propulsion Laboratory for providing us with their fiducial detection library.
This research was supported in part by National Science Foundation grants
ECS-0326095, IIS-0535282, IIS-1017134, CNS-0619937, IIS-0917318, CBET-0922784,
EECS-0926052, CNS-0960061, the DARPA program on Autonomous Robotic Manipulation,
the Army Research Office, the Okawa Foundation, the ATR Computational
Neuroscience Laboratories, and the Max-Planck-Society.\par}

 
{\small
\bibliographystyle{abbrvnat}
\bibliography{paper}
}

\end{document}